\documentclass[10pt,twocolumn,letterpaper]{article}

\usepackage{cvpr}              

\usepackage{graphicx}
\usepackage{amsmath}
\usepackage{amssymb}
\usepackage{booktabs}

\usepackage{tabularx}
\usepackage{float}

\usepackage[pagebackref,breaklinks,colorlinks]{hyperref}

\usepackage[capitalize]{cleveref}
\crefname{section}{Sec.}{Secs.}
\Crefname{section}{Section}{Sections}
\Crefname{table}{Table}{Tables}
\crefname{table}{Tab.}{Tabs.}

\def\cvprPaperID{5} 
\def\confName{CVPR}
\def\confYear{2023}

\begin{document}

\title{Fine-Grained Product Classification on Leaflet Advertisements}

\author{Daniel Ladwig$^1$ \hspace{1cm} Bianca Lamm$^{1,2}$ \hspace{1cm} Janis Keuper$^1$ \\
$^1$ IMLA, Offenburg University, $^2$ Markant Services International GmbH\\
{\tt\small dladwig@stud.hs-offenburg.de}, {\tt\small bianca.lamm@de.markant.com}, {\tt\small keuper@imla.ai}
}
\maketitle

\begin{abstract}
\noindent In this paper, we describe a first publicly available fine-grained product recognition dataset based on leaflet images. Using advertisement leaflets, collected over several years from different European retailers, we provide a total of 41.6k manually annotated product images in 832 classes. Further, we investigate three different approaches for this fine-grained product classification task, Classification by Image, by Text, as well as by Image and Text. The approach "Classification by Text" uses the text extracted directly from the leaflet product images. We show, that the combination of image and text as input improves the classification of visual difficult to distinguish products. The final model leads to an accuracy of $96.4\%$ with a Top-3 score of $99.2\%$.\\
\noindent \url{https://github.com/ladwigd/Leaflet-Product-Classification}\\

\end{abstract}

\section{Introduction}
\label{sec:intro}
\begin{figure}[!ht]
     \centering
     \def\svgwidth{\columnwidth}
\begingroup%
  \makeatletter%
  \providecommand\color[2][]{%
    \errmessage{(Inkscape) Color is used for the text in Inkscape, but the package 'color.sty' is not loaded}%
    \renewcommand\color[2][]{}%
  }%
  \providecommand\transparent[1]{%
    \errmessage{(Inkscape) Transparency is used (non-zero) for the text in Inkscape, but the package 'transparent.sty' is not loaded}%
    \renewcommand\transparent[1]{}%
  }%
  \providecommand\rotatebox[2]{#2}%
  \newcommand*\fsize{\dimexpr\f@size pt\relax}%
  \newcommand*\lineheight[1]{\fontsize{\fsize}{#1\fsize}\selectfont}%
  \ifx\svgwidth\undefined%
    \setlength{\unitlength}{450.98535252bp}%
    \ifx\svgscale\undefined%
      \relax%
    \else%
      \setlength{\unitlength}{\unitlength * \real{\svgscale}}%
    \fi%
  \else%
    \setlength{\unitlength}{\svgwidth}%
  \fi%
  \global\let\svgwidth\undefined%
  \global\let\svgscale\undefined%
  \makeatother%
  \begin{picture}(1,0.44480268)%
    \lineheight{1}%
    \setlength\tabcolsep{0pt}%
    \put(0,0){\includegraphics[width=\unitlength,page=1]{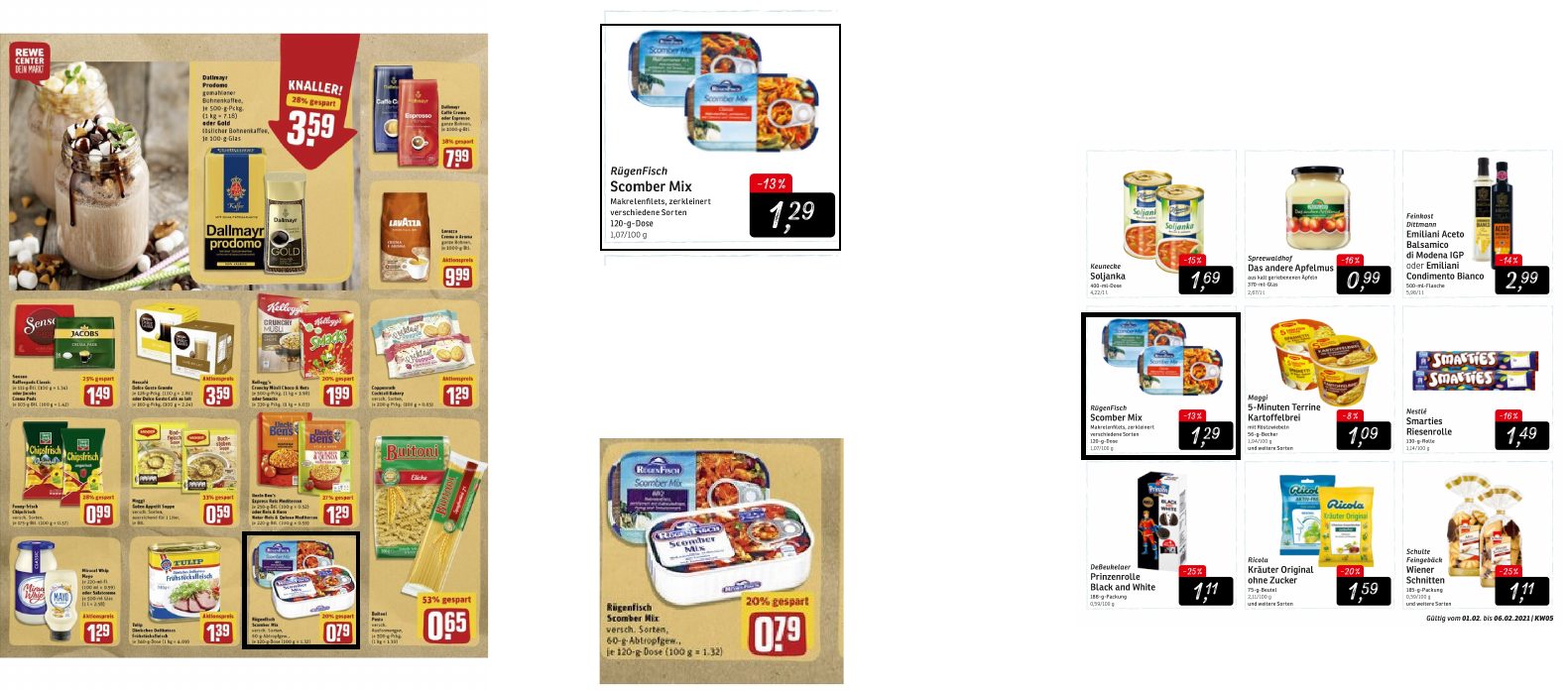}}%
    \put(0.35,0.21977225){\color[rgb]{0,0,0}\makebox(0,0)[lt]{\lineheight{1.25}\smash{\begin{tabular}[t]{c}{\scriptsize same}\\[-4pt]{\scriptsize product}\end{tabular}}}}%
    \put(0.4725,0.21977225){\color[rgb]{0,0,0}\makebox(0,0)[lt]{\lineheight{1.25}\smash{\begin{tabular}[t]{c}{\scriptsize different}\\[-4pt]{\scriptsize prices \& retailers}\end{tabular}}}}%
    \put(0,0){\includegraphics[width=\unitlength,page=2]{images/visual_abstract/visual_abstract.pdf}}%
  \end{picture}%
\endgroup%

    \caption{Price monitoring based on printed leaflets is a key data analysis task in retail, which technically can be defined as a fine-grained, multi-modal classification problem. We provide a first public dataset with 41.6k annotated samples for this task.}
    \label{fig:price_monitoring}
\end{figure}

\noindent The monitoring of product prices is an important data analysis task for retailers as their own price strategy heavily depends on the prices set by competitors. In this context, the monitoring of product advertisements in printed or online leaflets are the predominant source to obtain pricing and promotion activities from competitors. However, the highly unstructured and multi-modal (image + text information) nature of leaflets and the large number of often very similar products makes the underling product identification and matching task quite challenging. Figure\,\ref{fig:price_monitoring} depicts an example of the promotions of the same product in the leaflets of two different retailers.
From a Computer Vision perspective, the retail product price monitoring task resolves into several objectives, from product detection to fine-grained classification (FC). Due to its overall complexity, these tasks are currently mostly solved manually, requiring vast resources. In this work, we focus on the FC task and provide a first dataset with 41.6k manually annotated product images in 832 classes manually obtained from leaflets, alongside first baseline solutions. 

\section{Related Work}
\label{sec:related_work}
\noindent We focus our literature review on publicly available and annotated image collections for the application of \textit{fine-grained product classification} on images in a retail context. The data sources of the datasets vary strongly. First, images from web stores are used as a data source. \cite{elayanithottathil2021retail} introduced a {\it Retail Product Categorisation Dataset} which covers about 48k products with staged ”studio” product images in 21 categories. These images are recorded in controlled environments. The authors developed a concatenation of a Convolutional Neural Network (CNN) and a Long Short-Term Network for the FC task \cite{elayanithottathil2021retail}. For more realistic real-life scenarios product images "in the wild" are used. The {\it Products-10k} \cite{bai2020products} collection comprises about 10k product classes for about 150k "studio" images and "in the wild" images recorded by customers. The backbone of their approach is the model EfficientNet-B3 \cite{tan2019efficientnet}. Also, images of supermarket shelves serve as a data source. \cite{karlinsky2017fine} provides the {\it Retail-121} dataset consisting of 121 fine-grained retail product categories. For solving the FC task, the authors supply an own approach that is based on a non-parametric probabilistic model and a CNN \cite{karlinsky2017fine}. To the best of our knowledge, there are no datasets that are based on images from leaflets. Hence, we provide the first annotated dataset containing of product promotions cropped from leaflets.

\section{Dataset Description}
\label{sec:dataset_dscrptn}

\begin{figure}[t]
     \centering
     \begin{subfigure}[b]{\columnwidth}
         \centering
         \includegraphics[width=\columnwidth]{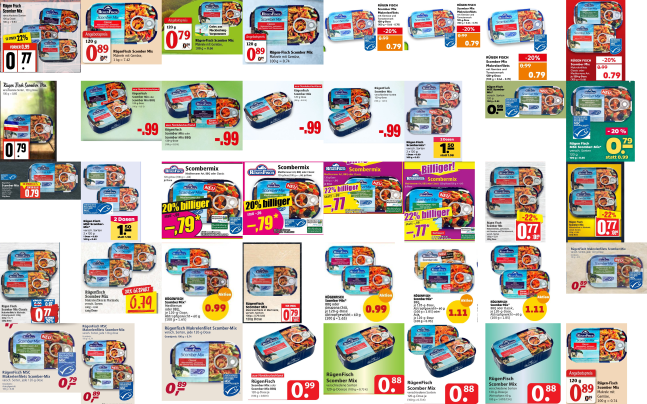}
     \end{subfigure}
     \par\bigskip
     \begin{subfigure}[b]{\columnwidth}
        \centering
    \includegraphics[width=0.8\columnwidth]{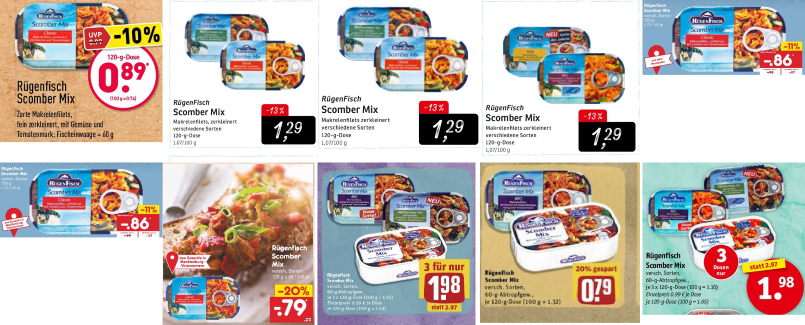}
    \end{subfigure}
    \caption{Training set (top) and test set (bottom) of a class.}
    \label{fig:train_test_set}
\end{figure}

\begin{figure}
     \centering
     \begin{subfigure}[b]{0.3\columnwidth}
         \centering
         \includegraphics[width=\textwidth]{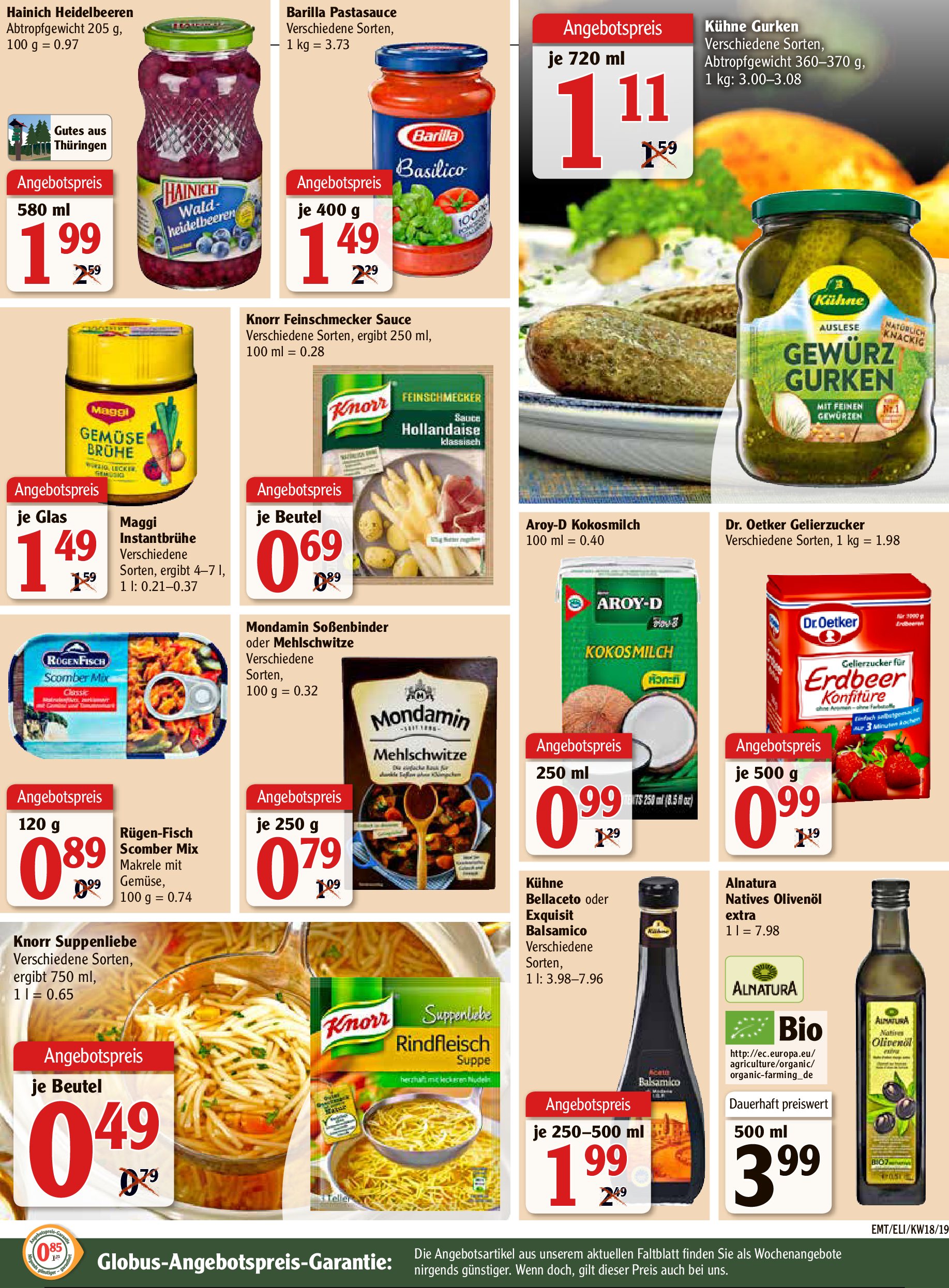} 
     \end{subfigure}
     \hfill
     \begin{subfigure}[b]{0.3\columnwidth}
         \centering
         \includegraphics[width=\textwidth]{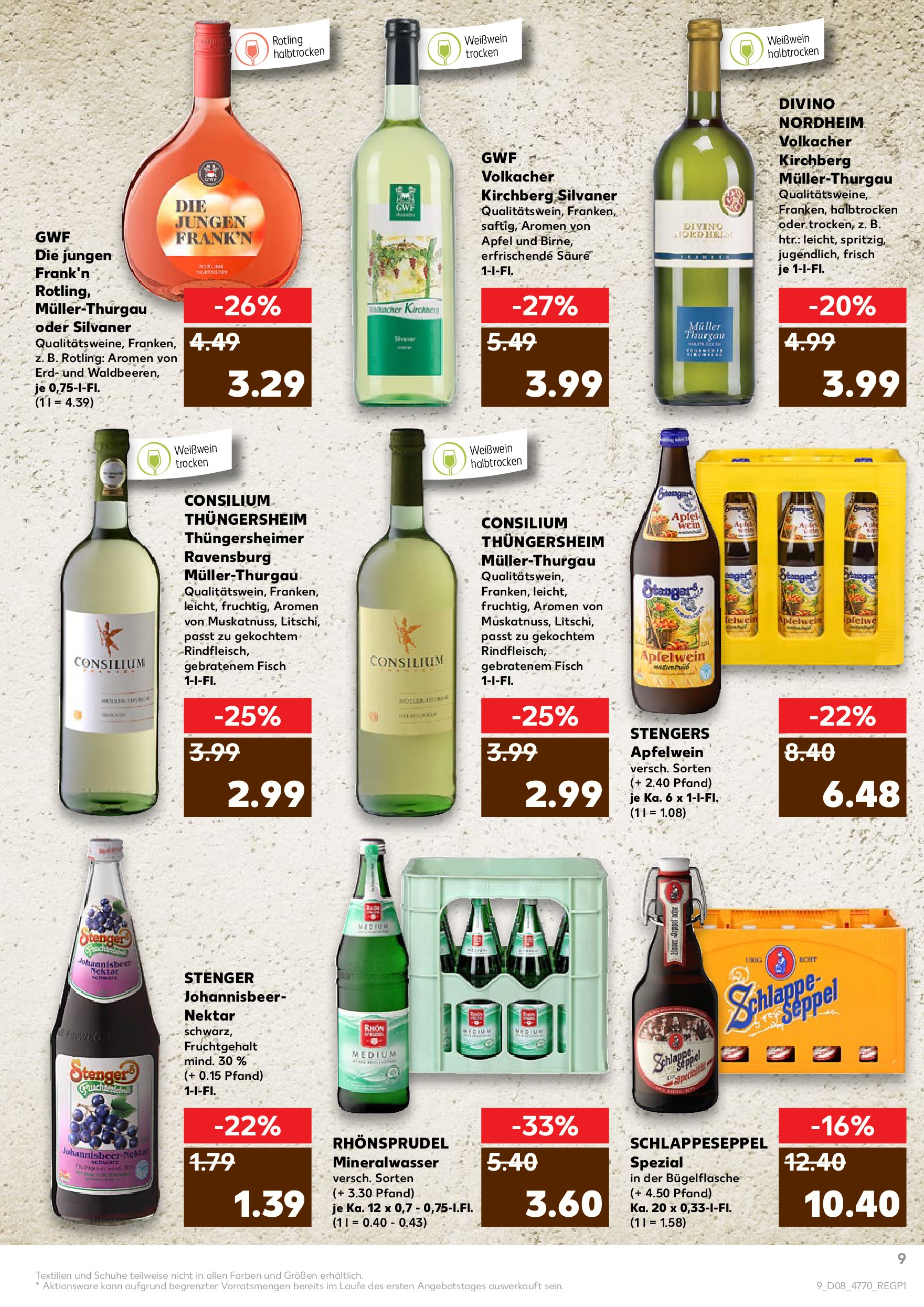}
     \end{subfigure}
     \hfill
     \begin{subfigure}[b]{0.3\columnwidth}
         \centering
         \includegraphics[width=\textwidth]{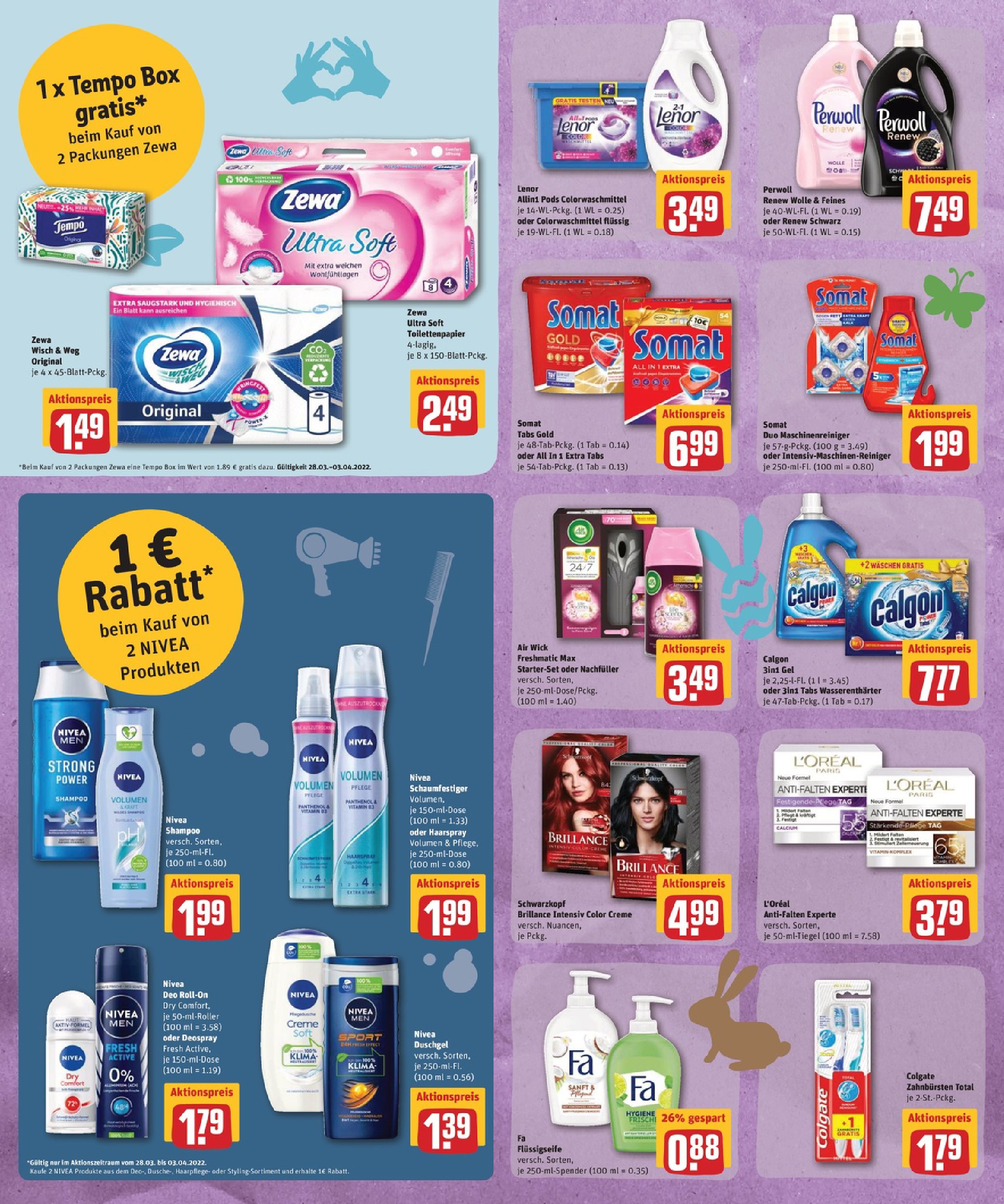}
     \end{subfigure}
        \caption{Three leaflet pages from different European retailers.}
        \label{fig:leaflet_examples}
\end{figure}

\noindent\textbf{Data Sources.} Our dataset is based on a large collection of full page images in JPG format provided by the company Markant Services International GmbH. The publicly available digital or manually scanned leaflets are circulated by well-known European retailers.
Figure\,\ref{fig:leaflet_examples} shows three representative samples. The leaflets were randomly collected from 132 different\footnote{Large retail chains often have different subsidiaries and brands which are potentially using the same product images for advertisement. In order to avoid poisoning of the test sets, we grouped all sub-brands into single retailers.} retail chains between calendar week 39 in 2016 to calendar week 38 in 2022. The leaflets advertise mainly food and beverages. But also non-food products like household goods, cosmetics, pet foods, or (small) electric devices are promoted. Each leaflet page has been manually segmented into product information boxes by humans. Each box must contain the product image, price, and description. Additional logos, price tags, or quality seals can also be contained. The cropped boxes from the leaflet pages form our provided image dataset. The original leaflets as well as the text information like prices, discounts, or product descriptions are not included in dataset.\\
\noindent\textbf{Dataset Properties.} Each class in the dataset represents one product. The dataset is composed of 832 classes and 41.6k images in total, split into a training set of 33,280 images and a test set of 8,320 images. Each class has 40 images in the training set and 10 images in the test set. All images adhere to a minimal resolution of at least 92 pixels in width and 138 pixels in height while the longer edge is always fixed to 512. This dataset has a size of 3.4 GB. Reducing the longer edge to a fixed value of 256, results in a dataset of a size of 1.2 GB. We provide both versions for public download. Figure\,\ref{fig:train_test_set} shows the training and the test set for a class. The images of a class are similar and diverse at the same time. There can be multiple images from one retailer in a split set but a retailer can only be represented in the training or in the test set for a class.

\section{Baseline Results}
\label{sec:methods_dcrptn}
\noindent We investigate baseline solutions of the FC task on our dataset by using different information of an image. The \textit{Classification by Image} uses the whole image as input information. The extracted text of an image is the input for the method \textit{Classification by Text}. Moreover, the combination of both information, \textit{Classification by Image and Text}, is analyzed. The following sections explain the details.

\subsection{Classification by Image.}
\begin{figure}
     \centering
     \begin{subfigure}[b]{0.47\columnwidth}
         \centering
         \includegraphics[width=\textwidth]{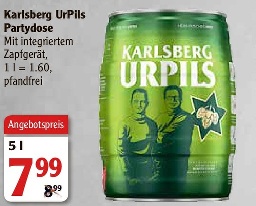}
         \label{fig:class_683_training_imgs}
     \end{subfigure}
     \hfill
     \begin{subfigure}[b]{0.47\columnwidth}
         \centering
         \includegraphics[width=\textwidth]{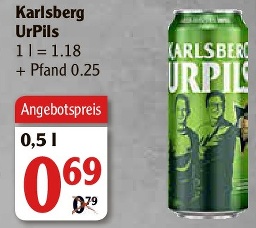}
         \label{fig:class_310_training_imgs}
     \end{subfigure}
        \caption{Illustration of two similar products. The left product is often confused with the right one by three of the four baselines.
        }
        \label{fig:misclassification_492_4}
\end{figure}

\noindent We provide first baseline results on our datasets, applying current state-of-the-art methods for FC. We used the dataset version with a longer edge length of 256.\\
\cite{wei2021fine} provides a review of current FC methods. For our baseline evaluation, we chose four different image classification models: {\it ResNet50} \cite{he2016deep}, {\it MobileNet V2} \cite{sandler2018mobilenetv2}, a {\it Vision Transformer} model \cite{dosovitskiy2020image} with a {\it vit$\_$b$\_$16} architecture and {\it ConvNeXt} \cite{liu2022convnet} using the base model architecture. Each model has been pretrained on {\it ImageNet} \cite{deng2009imagenet}. The following settings are used for the training process on a NVIDIA GeForce RTX 3090 GPU: 30 epochs of fine-tuning, a batch size of 32, and a SGD optimizer with a learning rate of 0.001 plus a momentum of 0.9. As Table\,\ref{tab:baseline_FC} shows, {\it MobileNet V2} gives the lowest test set accuracy result with an accuracy of $0.894$. However, this model needed the shortest training time. The {\it ResNet50} model and the {\it Vision Transformer} model have nearly the same test set accuracy of $0.907$ and $0.909$, respectively. Also the Top-5 accuracy scores of both models is about $0.971$. However, the {\it Vision Transformer} requires almost twice of the training time compared to  the {\it ResNet50} model. The best test set accuracy is reported for the {\it ConvNeXt} model with $0.921$, but its training took almost four hours. We use the \textit{ResNet50} model as the final image model because of its solid accuracy relative to an acceptable training time. In the fine-tuned model the last \textit{Fully Connected} layer is replaced with a \textit{Linear} layer of 2048-1024, followed by a \textit{ReLu} activation function and closing with a \textit{Linear} layer of 1024-832. The hyperparameter used are: batch size of 16 and optimizer momentum of 0.95. A torch ColorJitter is applied as data augmentation with a saturation of 0.5.\\
\textbf{Error Analysis.} 
By analysing falsely classified products, errors between visually similar classes occur. Figure\,\ref{fig:misclassification_492_4} exemplifies such products, which still match at a higher abstraction being very similar items from the same producer.
\begin{table}[]
    \centering
    \begin{tabularx}{\columnwidth}{c|X|X|X}
        model                       &test accuracy   &training time [h]  &Top-5 accuracy \\
        \hline
        {\it ConvNeXt}              &$0.921$    &$3.9$                &$0.974$\\
        {\it Vision Transformer}    &$0.909$    &$2.1$               &$0.971$\\
        {\it ResNet50}              &$0.907$    &$1.3$            &$0.971$\\
        {\it MobileNet V2}          &$0.894$    &$0.9$                &$0.964$\\
    \end{tabularx}
    \caption{Listing of the test accuracy scores, training time, and the Top-5 accuracy for tour FC dataset. The models {\it ConvNeXt, ResNet50, Vision Transformer} and {\it MobileNet V2} were trained.}
    \label{tab:baseline_FC}
\end{table}
\subsection{Classification by Text.}
\noindent The error analysis of the image model shows the difficulty to distinguish especially between products that are served in different package sizes or product variations that are not clearly visually represented on the product image itself. The leaflet product images mostly include a product description and the serving size. Those texts are not available in extracted digital form yet. To process them in further steps, they need to be extracted from the images first.\\
\textbf{OCR Extraction.}
The Tesseract OCR Engine \cite{smith2007overview} is chosen to extract texts from the images. This tool combines the text detection and the text recognition in one model. Different page segmentation modes (PSMs) \footnote{For further explanation: https://github.com/tesseract-ocr/tessdoc} can be set which alters the way text in the image is treated. Changing modes leads to different extracted text results. Since there is no digital truth text available to evaluate the quality of the extracted text, the accuracy of the classification problem is used to evaluate the quality of the extraction.\\
To extract much text out of different designed product images with changing background colors, text positions and text sizes, several image preprocessing steps and PSMs are combined. The dataset with the larger image sizes with a longer edge of 512 is used to extract the text to utilize the higher text resolution.\\
The extraction methods were developed incrementally. 
An extraction method was put together, the resulting wrong predicted images were inspected and a new fitting method for the poorly performing text extractions was developed.
In the final model the following methods were used to extract text from the varying input images.
OpenCV\cite{opencv_library} and its methods were utilized to import the images and to apply preprocessing. The first four methods use Tesseract with PSMs of 3, 6, 11, 12.
Another method applies a grayscaling with cvtColor and cv2.COLOR$\_$BGR2GRAY before applying Tesseract with the standard PSM of 3. 
For additional methods the gray color change was applied together with a resize of the image times four with Tesseract and PSMs 6 and 11. 
Finally a grayscaling, resize and cv2.threshold with cv2.THRESH$\_$OTSU, a thresh of 0, maxval 255 and Tesseract with PSM 11 is used.
The following preprocessing steps and model are used to evaluate the described methods. They resulted in between $73.6\%$ and $86.4\%$ accuracy.
\\
The combination of all developed extraction methods, to extract as much text from the image as possible, boosts the performance to $91.5\%$. Even if that means that there are duplicate texts in the collection. Extracting the 33,280 train images sequentially took 16.83h.\\
\noindent\textbf{Text preprocessing.}
The extracted texts include wrong letters and signs from noise detected as text in the image. The preprocessing has to be done carefully since package size descriptions can include OCR caused confusions of numbers and letters. In the final solution no stopword or special sign removal is done. The raw extracted texts which include duplicate text from combining the methods are encoded with scikit-learn TfidfVectorizer \cite{sklearn_TfidfVectorizer}.\\
\textbf{Text model.}
A linear SGDClassifier is trained on the tfidf encoded text and the loss function "modified$\_$huber" with an adaptive learning rate is used. Table\,\ref{tab:model_comination_results} shows the model accuracy is $91.5\%$ with a Top-5 score of $96.7\%$.
The text models performance is slightly worse than the image models but the correct predicted images differ. The following step combines those two models.

\subsection{Classification by Image and Text.}
\begin{figure}
    \centering
    \includegraphics[width=\columnwidth]{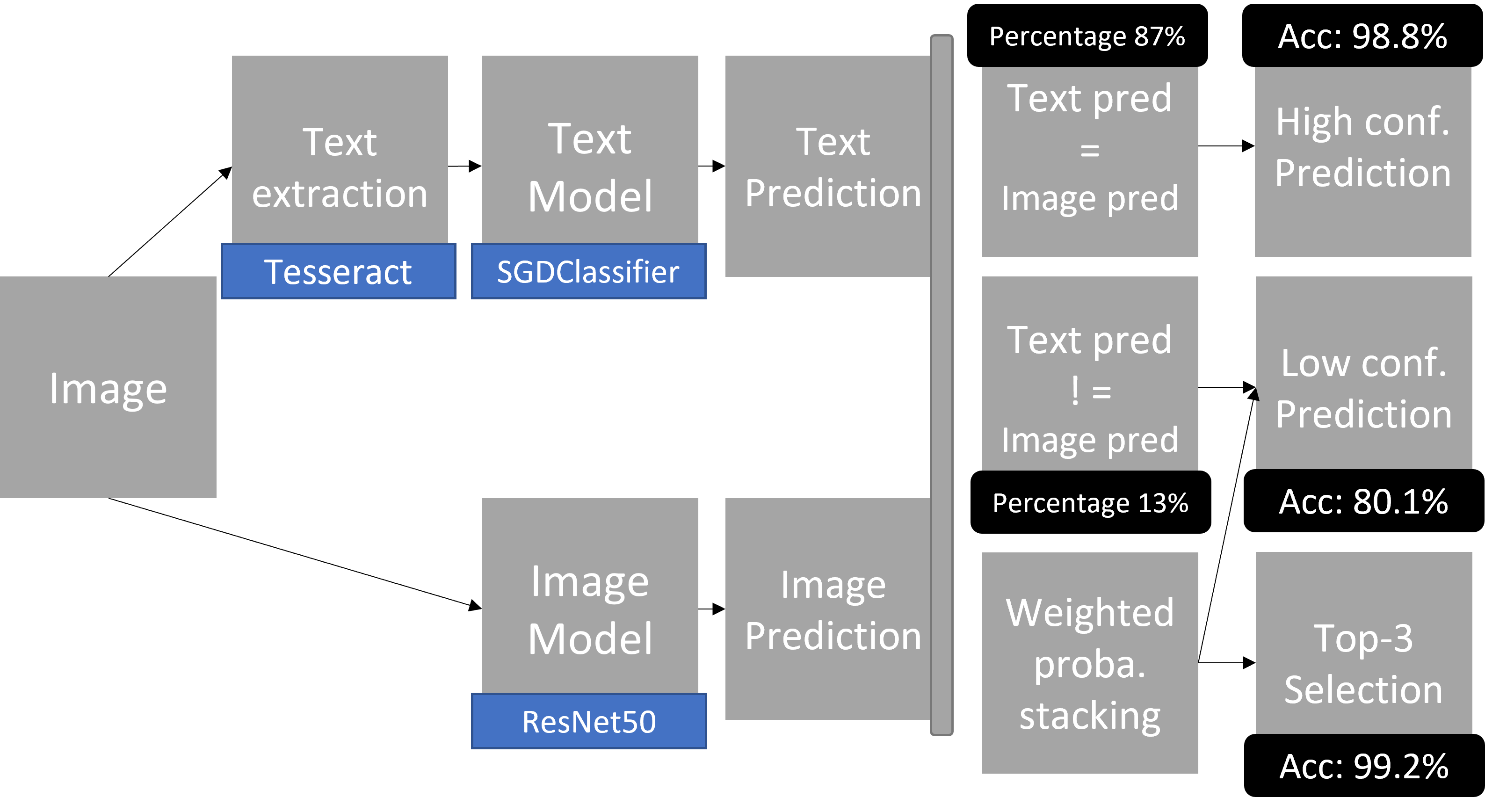}
    \caption{ Model Combination Process
    }
    \label{fig:model_combination}
\end{figure}

\noindent The first step to combine the text and image models is to apply a softmax to each probability predictions. After that a weighted probability stacking is done by giving the text model a higher weight, as the SGDClassifier predicts way lower decision probabilities than the ResNet50 model. Weighting the text model allows to make an impact after stacking the probabilities. The final predictions are differentiated in predictions where both models agree (labeled as high confidence prediction) and predictions where they do not (labeled as low confidence prediction). This allows to separate those low confidence predictions to be potentially reviewed manually with the help of a Top-3 selection with a very high accuracy of $99.2\%$. Figure\,\ref{fig:model_combination} displays the process of combining the models and the resulting predictions.

\begin{figure}
    \centering
    \includegraphics[width=\columnwidth]{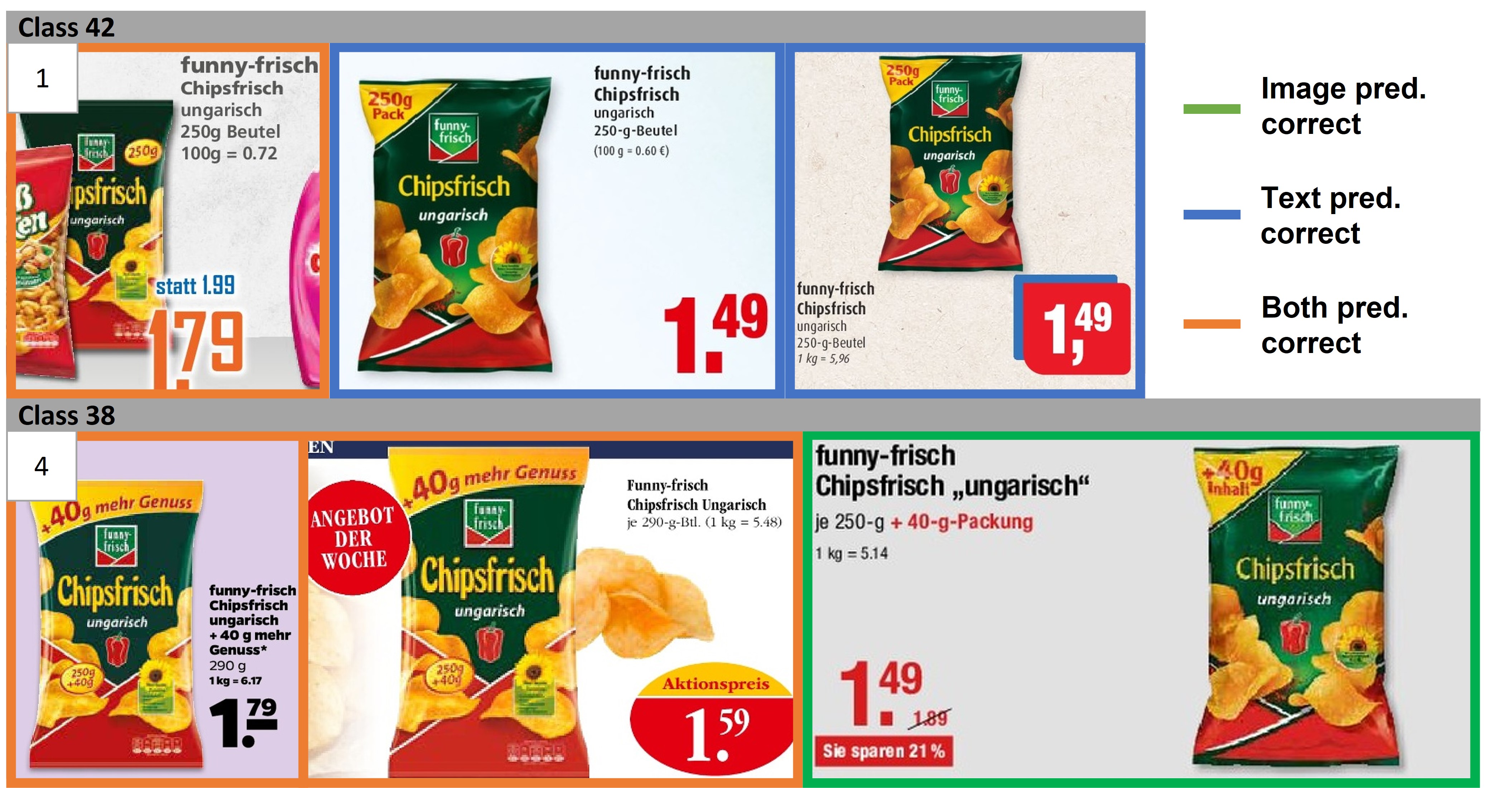}
    \caption{ Model Confusion Example 
    }
    \label{fig:model_confusions}
\end{figure}

\noindent\textbf{Combination Results.}
The prediction stacking method by combining the probabilities of the text and image model results in an overall accuracy of $96.4\%$ which is a significant increase from the models alone.
Interesting is the calculated Top-3 accuracy of $99.2\%$. This shows the ability of the combined models to find the correct class in the fine-grained problem and rate it high.
The gap between the prediction and Top-3 selection displays the difficulty to make the correct choice when the products do not differ much, sometimes only one attribute differentiates them. There is still space for improvement in the combination process of the two models. The accuracy that the image model prediction or the text model prediction is correct lays by $98.0\%$ which is not quite reached by the used weighted probability stacking method.
\begin{table}
	\centering
	\begin{tabularx}{\columnwidth}{cXccc}
		\toprule
		Model & Method & Accuracy & Top-3 & Top-5\\
		\midrule
		Final ResNet50 & Image & $0.925$ & $0.961$ & $0.969$\\
		SGDClassifier & Text & 0$.915$ & $0.962$ & $0.967$\\
		Combined & Weighted proba. & $0.964$ & $0.992$ & $0.993$\\
		\bottomrule
	\end{tabularx}
	\caption{Model and Combination Results}
	\label{tab:model_comination_results}
\end{table}
\\
\textbf{Error Analysis.} A manual review of the wrong predicted images and classes of the final combined model revealed the following findings. As in the beginning presumed the products that are served in different variations or package sizes can cause confusions for the final prediction. The high Top-5 score shows that the correct product can be found but the final choice which variation is presented is still a difficult task. Figure\,\ref{fig:misclassification_492_4} already shows an example of two similar product servings. Figure\,\ref{fig:model_confusions} shows two classes of products that are distinguished by serving size. The Top-3 products are served in 250g and the bottom products are served in 290g. The package size displayed on the product varies, even within the class. This explains the confusion between those classes for the image model. The text model could possibly distinguish those two from the given package size. Although the last image labels the package size as "250g + 40g" instead of the expected 290g which can lead to confusions. In this case the combined model could predict both classes correctly, but this is an example how difficult the differentiation between similar classes can be. 
Changing product descriptions or difficult text extractions impact the text model predictions negatively.

\section{Conclusion}
\noindent In this paper, we provide a dataset and present a baseline solution for fine-grained image classification of retail products on leaflets.
We propose methods to optimize the product classification by combining image classification with text classification based on text directly extracted from the  images.
The presented methods allow to extract and process text from varying product images.
Improved results compared to the image classification alone support the idea to make use of existing text.
Combining text and image classification improves the ability to distinguish between visually similar product representations.
As a first part of the price monitoring task we create a sufficient baseline for product matching. Future work will be the extracting of the prices of the product promotions.
This is challenging due to the huge variance in color, size, and position in the promotion.
\newpage

{\small
\bibliographystyle{ieee_fullname}
\bibliography{egbib}
}

\end{document}